\documentclass[runningheads]{llncs}

\usepackage{eccv}
\usepackage{eccvabbrv}
\usepackage{graphicx}
\usepackage{booktabs}
\usepackage[accsupp]{axessibility}
\usepackage{amsmath,amssymb}
\usepackage{algorithm}
\usepackage{algpseudocode}
\usepackage{tikz}
\usetikzlibrary{arrows.meta,positioning,fit}
\usepackage{placeins}
\usepackage{hyperref}

\begin{document}

\title{Cache-Aware Conv3D Lowering Across Embedded World-Model Decoders}
\titlerunning{Cache-Aware Conv3D Lowering Across Embedded World-Model Decoders}

\author{
Jiaming Zhang\inst{1,2}
\and
Wu Yang\inst{2}
\and
Shuai Tao\inst{2}
\and
Wulong Liu\inst{2}
}

\authorrunning{J.~Zhang et al.}

\institute{
University of Wisconsin--Madison, Madison, WI, USA
\and
Beta Infinity, Beijing, China
}

\maketitle

\begin{abstract}
Generative world models can provide visual rollouts for embodied planning, yet
their feasibility on edge devices depends not only on the learned model but
also on how the execution runtime represents its operations. We introduce a
cache-aware lowering that expresses supported causal Conv3D calls as batched
spatial Conv2D operations while preserving pretrained weights, temporal-cache
semantics, convolution parameters, bias placement, and output layout. Across
the complete Cosmos3-Edge image-to-video pipeline on a 64-GB NVIDIA Jetson AGX
Orin, the proposed route accelerates VAE decoding by approximately $7\times$
and reduces complete-generation latency by more than $2\times$, while repeated
decoder evaluations maintain complete fast-path coverage without fallbacks.
The unchanged lowering also improves Cosmos3-Nano and transfers to
LingBot-World's architecturally distinct Wan2.1 VAE. A clean-device comparison
against fully specialized TensorRT shows that TensorRT provides a further
$1.36\times$ steady-state improvement, but requires substantially greater
per-module and per-runtime-state AOT specialization. Same-latent BF16 and FP32
evaluations characterize the finite-precision differences introduced by the
alternative execution order. Together, these results position cache-aware
lowering as a lightweight runtime optimization that recovers most of the
available decoder acceleration without modifying the learned models
themselves.

\keywords{World models \and Video VAE \and Edge inference
\and Runtime systems \and GPU optimization}

\end{abstract}

\section{Introduction}
\label{sec:introduction}

Learned world models support control by enabling an agent to reason through
imagined observations or latent trajectories before acting
~\cite{ha2018worldmodels,finn2016deep,hafner2019planet,
hafner2020dreamer}. Foundation-scale systems extend this concept toward
general-purpose video generation and Physical AI models
~\cite{cosmos2025platform,cosmos3release,wan2025}. Because a planner may need
to evaluate several possible futures before selecting an action, rollout
latency directly limits both deliberation speed and the number of alternatives
that can be considered. This systems-level concern is separate from whether
the generated video is sufficiently accurate for closed-loop control.

We study local execution on a 64-GB NVIDIA Jetson AGX Orin, whose unified-memory
architecture provides substantially less compute capacity and memory bandwidth
than datacenter GPUs. Robotic systems may alternatively rely on cloud or fog
offloading~\cite{tahir2025edge,ichnowski2022fogros2}, but such approaches
introduce additional networking, availability, and latency considerations.

The central question of this work is whether a fixed pretrained causal video
decoder can be executed more efficiently through a runtime-only
transformation. We do not retrain, quantize, prune, distill, replace, or
otherwise modify the learned model. A Cosmos3-Nano profile attributes 139.91
of 145.82 VAE-decode GPU seconds, or 95.95\%, to baseline Conv3D, with the
largest costs concentrated in calls with short temporal extents and large
spatial feature maps. To address this pattern, we lower each supported
temporal kernel tap to batched spatial Conv2D while preserving the pretrained
weights, causal-cache ordering, convolution parameters, bias placement, and
output layout. Unsupported calls return to the original Conv3D implementation,
while route counters expose unsupported cases and fallbacks.

We evaluate Cosmos3-Edge as the primary deployment case through its complete
Diffusers-native image-to-video pipeline. Edge uses the same byte-identical
Wan2.2 VAE checkpoint as Cosmos3-Nano but exercises it through a different
conditioning and generation path. We separately measure image-conditioning VAE
encoding, denoising and other pipeline operations, and final VAE decoding,
while retaining the Nano profile and complete text-to-video sweep. Finally, we
apply the unchanged lowering to
LingBot-World~\cite{robbyantteam2026advancingopensourceworldmodels}, whose
distinct Wan2.1 VAE provides evidence of transfer across decoder
architectures. This evaluation design distinguishes isolated decoder
acceleration from improvements that remain visible at the complete-generation
boundary across different workloads, frame lengths, and VAE configurations.

\subsection*{Our Contributions}

\noindent\textbf{(1) Edge deployment characterization.}
We connect the reconciled Cosmos3-Nano Conv3D profile to a controlled
Cosmos3-Edge deployment using the byte-identical Wan2.2 VAE. We separate
complete I2V latency into encoder, decoder, and unchanged
transformer-and-other work.

\noindent\textbf{(2) Cache-aware lowering.}
We express supported cached causal Conv3D calls as batched Conv2D operations
while preserving temporal indexing and model semantics. The route includes
guarded fallback, reversible installation, derived-weight caching, and explicit
route accounting.

\noindent\textbf{(3) Cross-boundary and cross-decoder evaluation.}
We report repeated Edge generation and decoder measurements, retain the
Cosmos3-Nano generation sweep, and test decoder-only transfer on
LingBot-World's distinct Wan2.1 VAE. We additionally report numerical
agreement, route coverage, completion, and memory behavior.

\noindent\textbf{(4) Production-runtime comparison.}
We construct a full-coverage TensorRT baseline for the 17-frame Edge decoder.
TensorRT achieves the lowest steady-state latency under aggressive fixed-model
AOT specialization, while our generic runtime lowering captures 95.55\% of
the absolute eager-to-TensorRT latency reduction without per-module or
per-state engine construction.

\section{Related Work}
\label{sec:related}

\paragraph{World Models and Video Generation.}
World models learn predictive dynamics that allow an agent to reason through
imagined observations or latent trajectories
~\cite{ha2018worldmodels,hafner2019planet,hafner2020dreamer}. Visual foresight
makes this connection more explicit by combining action-conditioned video
prediction with model-predictive control~\cite{finn2016deep}. More recent
systems extend predictive modeling toward general-purpose video generation and
Physical AI. Cosmos provides pretrained world foundation models
~\cite{cosmos2025platform,cosmos3release}, while Wan introduces an open family
of large-scale video generators~\cite{wan2025}. LingBot-World further develops
an interactive world model using the Wan video stack
~\cite{robbyantteam2026advancingopensourceworldmodels}. These systems motivate
the need for efficient rollout generation. However, our study does not compare
predictive accuracy, interaction quality, or policy performance. Instead, we
examine whether the execution of fixed pretrained video VAEs can be accelerated
through a runtime transformation alone. This distinction allows the learned
behavior of the model to remain unchanged while isolating the effect of the
underlying execution strategy.

\paragraph{Spatiotemporal Computation and Decoder Acceleration.}
Previous work has restructured 3D convolution by separating its spatial and
temporal computation. P3D and R(2+1)D replace or factorize 3D kernels into
combinations of spatial and temporal operations for network design and
optimization~\cite{qiu2017p3d,tran2018closer}. Our approach similarly exposes
the spatial computation within Conv3D, but it does not introduce a learned
factorization or alter the architecture. Instead, each temporal tap of an
existing causal Conv3D is evaluated as spatial Conv2D over the valid cached
frames, while the original weights and temporal indexing are maintained.
Unlike video-VAE approaches that redesign, prune, replace, or distill decoder
components~\cite{zou2025turbovaed,zhu2026flashvaed}, our lowering changes only
the runtime representation of supported calls inside fixed pretrained
decoders. Therefore, the method targets execution efficiency without requiring
additional training data, parameter updates, or changes to the decoder's
learned function.

\paragraph{Embedded Runtimes, Compilation, and Serving.}
Embedded robotic systems operate under strict compute, memory, and latency
constraints~\cite{tahir2025edge,ichnowski2022fogros2}. TVM and Ansor provide
graph-, operator-, and hardware-aware tensor transformations
~\cite{chen2018tvm,zheng2020ansor}, while Triton enables developers to
construct programmable GPU kernels~\cite{tillet2019triton}. vLLM-Omni provides
stage-oriented execution for multimodal and diffusion pipelines
~\cite{yin2026vllmomni}. Our work addresses a complementary operator-level
problem inside the decoder stage: preserving temporal-cache semantics while
selecting a guarded runtime route. For this reason, we evaluate support
conditions, fallback behavior, route coverage, numerical agreement, memory
use, and complete-generation latency together. We additionally compare against
a full-coverage TensorRT deployment as a production-runtime reference.
TensorRT achieves higher throughput on the evaluated 17-frame workload. We
therefore do not claim universal superiority over optimized Conv3D or global
optimality across runtimes and hardware.

\section{Deployment Characterization}
\label{sec:characterization}

\subsection{Cosmos3-Nano Profile and Edge Checkpoint Identity}

We profile a 17-frame, five-step, $1280\times720$ Cosmos3-Nano request using
Nsight Systems 2026.3.1. Conv3D contributes 139.91 of 145.82~s of VAE GPU
time, corresponding to 95.95\%. In addition, one implicit BF16
3D-convolution kernel family accounts for 75.6\% of the captured kernel time.
Within the measured 17-frame decoder boundary, the target workload contains
169 causal Conv3D invocations. The highest-cost signatures are summarized in
\cref{tab:signatures}; the most expensive calls combine short temporal extents
with large spatial feature maps, motivating the temporal-tap lowering introduced
in \cref{sec:lowering}.

The Cosmos3-Edge VAE is byte-identical to the Nano Wan2.2 VAE. Both the
checkpoint bytes and all 196 parameter tensors match exactly. Therefore, the
Nano profile identifies the same operator family targeted in the Edge
deployment. However, the 95.95\% attribution belongs specifically to the
profiled Nano execution and should not be interpreted as an independently
measured Edge-specific profile.

\begin{table}[!htbp]
\caption{Highest-cost causal Conv3D signatures. Cosmos3-Nano uses Nsight GPU
time at $1280\times720$; LingBot-World uses CUDA events at $480\times832$.
Edge uses the byte-identical Wan2.2 VAE but is not independently profiled.}
\label{tab:signatures}
\centering
\scriptsize
\setlength{\tabcolsep}{3.2pt}
\begin{tabular*}{\textwidth}{@{\extracolsep{\fill}}llcccrrr@{}}
\toprule
Model & Rank & $C_{in}\!\to\!C_{out}$ & $T_{in}\!\to\!T_{out}$ &
$H\!\times\!W$ & $T_c$ & Calls & Time / VAE share \\
\midrule
Nano
& 1 & 256$\to$256 & 4$\to$4 & 360$\times$640 & 2 & 15
& 26.08~s / 17.88\% \\
Nano
& 2 & 512$\to$512 & 4$\to$4 & 180$\times$320 & 2 & 15
& 25.59~s / 17.55\% \\
Nano
& 3 & 1024$\to$1024 & 2$\to$2 & 90$\times$160 & 2 & 15
& 12.76~s / 8.75\% \\
\midrule
LingBot
& 1 & 96$\to$96 & 4$\to$4 & 480$\times$832 & 2 & 18
& 12.620~s / 27.0\% \\
LingBot
& 2 & 192$\to$192 & 4$\to$4 & 240$\times$416 & 2 & 18
& 12.343~s / 26.4\% \\
LingBot
& 3 & 192$\to$192 & 4$\to$4 & 240$\times$416 & 1 & 6
& 5.115~s / 10.9\% \\
\bottomrule
\end{tabular*}
\end{table}

\subsection{Distinct Wan2.1 Transfer Profile}

At 17 frames, LingBot executes 164 Conv3D calls across 28 signatures. The
summed CUDA-event time for these calls is 45.801~s, or 97.8\% of the
46.812~s clean VAE-decode time. The instrumented decode takes 48.786~s,
including approximately 1.97~s of measurement overhead. A separate decode
with synchronization after every call takes 46.54~s under exclusive GPU
access.

All observed calls use five-dimensional tensors, unit temporal stride, one
convolution group, and a temporal kernel size of three. The two highest-cost
signatures contribute 53.4\% of the attributed Conv3D time. Computation is
therefore again concentrated in calls with short temporal kernels and large
spatial feature maps. Although LingBot-World uses a distinct Wan2.1 VAE, its
dominant calls satisfy the same execution contract targeted by the proposed
lowering.

\section{Cache-Aware Runtime Lowering}
\label{sec:lowering}

\subsection{Temporal-Tap Conv2D Formulation}

Let $Z$ represent the cache-augmented input and let $W$ represent the unchanged
Conv3D weights:
\[
\begin{aligned}
Z &\in \mathbb{R}^{N\times C_{\mathrm{in}}\times T_{\mathrm{in}}
\times H_{\mathrm{in}}\times W_{\mathrm{in}}},\\
W &\in \mathbb{R}^{C_{\mathrm{out}}\times C_{\mathrm{in}}
\times K_t\times K_h\times K_w}.
\end{aligned}
\]
For unit temporal stride and one convolution group, temporal tap $k$ selects
the following source time:
\begin{equation}
\label{eq:source-time}
q_k(t)=t+k d_t-p_l,
\end{equation}
where temporal positions outside the valid input range contribute zero. With
$W_k=W[:,:,k,:,:]$, PyTorch's cross-correlation convention gives
\begin{equation}
\label{eq:conv3d}
Y_{n,:,t,:,:}
=
b+\sum_{k=0}^{K_t-1}
\operatorname{Conv2D}
\left(
Z_{n,:,q_k(t),:,:},
W_k;
\mathbf{s},\mathbf{p},\mathbf{d}
\right),
\end{equation}
where $\mathbf{s}=(s_h,s_w)$, $\mathbf{p}=(p_h,p_w)$, and
$\mathbf{d}=(d_h,d_w)$ denote the original spatial stride, padding, and
dilation parameters, respectively. Therefore, the unchanged Conv3D kernel can
be reconstructed by summing the spatial response from each temporal tap and
adding the bias exactly once.

\subsection{Temporal Intervals}

Let $C$ denote the cache, $X$ denote the current input, and $Z=[C;X]_T$ denote
their concatenation along the temporal dimension. The effective left padding
is $p_l=p_l^{\mathrm{base}}-|C|_T$, which produces
\begin{equation}
\label{eq:output-time}
T_o=
\left\lfloor
\frac{|Z|_T+p_l+p_r-d_t(K_t-1)-1}{s_t}
\right\rfloor+1.
\end{equation}
For the supported calls, $s_t=1$. Defining
$\delta_k=k d_t-p_l$, the intersection of the constraints
$0\leq o<T_o$ and $0\leq o+\delta_k<|Z|_T$ gives
\begin{equation}
\label{eq:interval}
o_l=\max(0,-\delta_k),\qquad
o_h=\min(T_o,|Z|_T-\delta_k).
\end{equation}
The corresponding source interval is
\[
[z_l,z_h)=[o_l+\delta_k,o_h+\delta_k).
\]
Both intervals have an equal length of $o_h-o_l$. This equality enables the
valid temporal range to be folded into the batch dimension and evaluated with
one Conv2D call for each active temporal tap. As a result, the implementation
avoids evaluating temporally invalid frames and preserves the same causal
alignment produced by the original cached Conv3D operation. The interval
construction also remains valid for temporal dilation because $d_t$ is already
included in the tap offset $\delta_k$.

\subsection{Guarded Runtime Procedure}

\begin{figure}[!htbp]
\centering
\begin{tikzpicture}[
  node distance=3.5mm and 4mm,
  box/.style={
    draw,
    rounded corners,
    align=center,
    minimum height=7mm,
    text width=0.18\linewidth
  },
  arr/.style={-{Latex[length=1.6mm]},thick}
]
\node[box] (cache)
  {Concatenate causal cache and input};

\node[box, right=of cache] (interval)
  {Find valid output interval for tap $k$};

\node[box, right=of interval] (fold)
  {Fold interval time into batch};

\node[box, right=of fold] (conv)
  {One batched Conv2D with $W_k$};

\node[box, below=of fold] (restore)
  {Restore time, accumulate taps, add bias once};

\draw[arr] (cache) -- (interval);
\draw[arr] (interval) -- (fold);
\draw[arr] (fold) -- (conv);
\draw[arr] (conv.south) |- (restore.east);
\draw[arr] (restore.west) -| (interval.south);

\node[
  draw,
  dashed,
  fit=(interval)(fold)(conv)(restore),
  inner sep=2mm,
  label={below:temporal-tap loop}
] {};
\end{tikzpicture}

\caption{Cache-aware lowering of each temporal tap to one batched Conv2D.}
\label{fig:lowering-overview}
\end{figure}
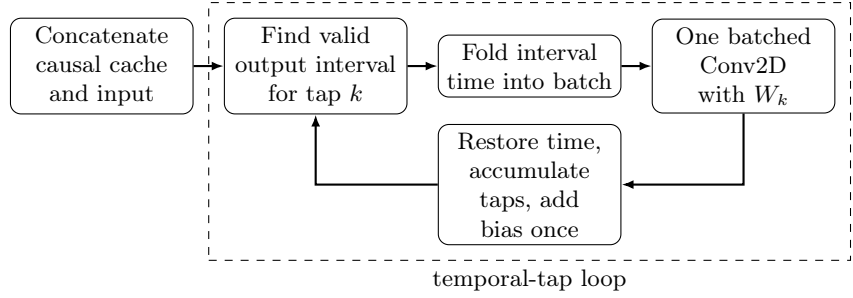

\Cref{fig:lowering-overview} summarizes the temporal-tap data flow, while
\cref{alg:lowering} specifies the corresponding validation, fallback, and
accumulation procedure. Unsupported calls return to the original Conv3D route,
and strict mode terminates validation when an unsupported case or execution
error occurs.

\begin{algorithm}[!htbp]
\caption{Cache-aware Conv3D-to-Conv2D lowering}
\label{alg:lowering}
\begin{algorithmic}[1]
\Require $X,C,W,b,\boldsymbol{p},\boldsymbol{s},\boldsymbol{d},g$
\Statex \textbf{State:} strict flag and route counters $\rho$

\Statex \textbf{\(\triangleright\) Validate and prepare}
\State increment $\rho.\mathrm{calls}$
\State \textbf{if} $X$ or $W$ is not a 5-D tensor, $s_t\neq1$, or $g\neq1$:
  \Return \textsc{Fallback}
\State $Z\gets[C;X]_T$ if cached, else $X$; compute $p_l,p_r,T_o$
\State \textbf{if} $p_r<0$ or $T_o<1$: \Return \textsc{Fallback}
\State $A\gets\varnothing$

\Statex \textbf{\(\triangleright\) Lower temporal taps}
\For{$k=0,\ldots,K_t-1$}
  \State compute $o_l,o_h,z_l,z_h$ using \cref{eq:interval}
  \State \textbf{if} $o_l\geq o_h$: \textbf{continue}
  \State $W_k\gets\operatorname{cached\_contiguous}(W[:,:,k,:,:])$
  \State $U_2\gets\operatorname{fold}_{NT}
    (Z[:,:,z_l:z_h,:,:])$
  \State $R\gets\operatorname{unfold}_{T}\!\left(
    \operatorname{Conv2D}
    (U_2,W_k;\mathbf{s},\mathbf{p},\mathbf{d})\right)$
  \State \textbf{if} $A=\varnothing$:
    $A\gets\operatorname{zeros}
    (N,T_o,C_{\mathrm{out}},H_o,W_o)$
  \State $A[:,o_l:o_h,:,:,:]\mathrel{+}=R$
\EndFor

\Statex \textbf{\(\triangleright\) Finalize}
\State \textbf{if} $A=\varnothing$: \Return \textsc{Fallback}
\State add $b$ once; restore layout; increment $\rho.\mathrm{fast}$
\State \Return contiguous output
\Statex \textbf{Exception:} re-raise if strict; otherwise fallback
\end{algorithmic}
\end{algorithm}

\FloatBarrier

\section{Experimental Methodology}
\label{sec:methodology}

\subsection{Systems, Boundaries, and Provenance}

\Cref{tab:systems} summarizes the evaluated systems. Cosmos3-Edge is the
primary deployment case, Cosmos3-Nano provides the detailed profile and
text-to-video sweep, and LingBot-World provides the cross-architecture
transfer evaluation.

\begin{table}[!ht]
\caption{Evaluated systems. Compression lists spatial/temporal factors.}
\label{tab:systems}
\centering
\scriptsize
\setlength{\tabcolsep}{3pt}
\begin{tabular*}{\textwidth}{@{\extracolsep{\fill}}lccclc@{}}
\toprule
Association & VAE & $z$ & Compression & Relation & Boundary \\
\midrule
Cosmos3-Edge
& Wan2.2 TI2V & 48 & $16\times/4\times$
& Primary deployment & Full I2V + decoder \\
Cosmos3-Nano
& Wan2.2 TI2V & 48 & $16\times/4\times$
& Byte-identical VAE & Full T2V + decoder \\
LingBot-World
& Wan2.1 & 16 & $8\times/4\times$
& Distinct architecture & Decoder only \\
\bottomrule
\end{tabular*}
\end{table}

Model artifacts are loaded offline from recorded repository snapshots, with
identifiers, revisions, configurations, source dtypes, and checkpoint SHA-256
values recorded for provenance. All experiments run on one 64-GB NVIDIA Jetson
AGX Orin with Python 3.12.13, PyTorch 2.11.0+cu130, CUDA 13.0, and cuDNN
9.19.0. Cosmos3-Nano uses Diffusers 0.38.0, while the frozen Cosmos3-Edge
environment uses Diffusers 0.40.0.dev0 at commit \texttt{b48d49d8}. The
production-runtime comparison adds TensorRT 10.16.2.10 while retaining the
same PyTorch, CUDA, cuDNN, and Diffusers stack. Primary latency measurements
use BF16, with FP32 reserved for numerical validation. The proposed route uses
PyTorch Conv2D; TensorRT is evaluated separately as a production-runtime
baseline. Quantization, pruning, decoder replacement, and model parallelism
are not used.

\subsection{Complete-Pipeline Workloads}

Cosmos3-Edge is evaluated through the Diffusers-native image-to-video path
using a fixed conditioning image and prompt at $480\times832$, 50 denoising
steps, guidance 5.0, seed 0, and BF16 execution. We evaluate 25 frames as the
primary deployment workload and five frames as the minimum runtime-accepted
request producing a nondegenerate temporal sequence. With $4\times$ temporal
compression, requests below five frames produce only one output frame.

The two configurations place different fractions of complete-generation time
inside the VAE and test the expected workload-share dependence. Frame-length
scaling is evaluated separately through the decoder-only sweep at 5, 9, 17,
33, and 49 frames.

Each complete-pipeline route uses three fresh-container repetitions under the
same warm-up procedure, and we report the median. We synchronize execution and
measure $T_{\mathrm{e2e}}$, image-conditioning $T_{\mathrm{encode}}$, final
$T_{\mathrm{decode}}$, and
\[
T_{\mathrm{dit+other}}=
T_{\mathrm{e2e}}-T_{\mathrm{encode}}-T_{\mathrm{decode}}.
\]
Calls are attributed by the active \texttt{vae.encode} or
\texttt{vae.decode} runtime scope.

The retained Nano sweep requests $\{5,9,17,33,49\}$ frames at
$1280\times720$, using five denoising steps, 24 FPS, seed 123, guidance 6.0,
flow shift 10.0, and text-to-video mode. Its synchronized
\texttt{omni.generate} timer includes tokenization, text encoding, denoising,
and VAE decoding but excludes host conversion, MP4 encoding, and file writing.
Each frame-count/route pair contains one fresh-process measurement after one
excluded warm-up.

\subsection{Decoder Sweeps and Acceptance}

For the Edge decoder evaluation, deterministic BF16 latents are generated for
each frame count and shared between baseline and optimized routes. We decode
5, 9, 17, 33, and 49 output frames using three fresh-process repetitions per
route and frame count, each with one untimed warm-up followed by one
CUDA-synchronized timed decode. For the tested $4k+1$ frame counts, the latent
shape is $[1,48,(F-1)/4+1,30,52]$.

LingBot uses synthetic decoder-compatible BF16 latents with batch size 1,
16 channels, temporal extent $T$, and spatial size $60\times104$. We use one
fresh-process measurement per route and frame count. These tests establish
decoder arithmetic and runtime transfer rather than complete LingBot generation
speedup or perceptual quality.

A point is accepted only if execution succeeds, returns a fresh nonempty
tensor of the expected shape, introduces no unexpected NaN or Inf values, and
reports consistent route counters. Baselines must contain no fast-path hits,
while optimized runs must use the intended route without unexplained
fallbacks.

\subsection{Clean Production-Runtime Comparison}
\label{sec:runtime-methodology}

We separately compare native PyTorch, our lowering, and full-coverage
TensorRT on the 17-frame Cosmos3-Edge decoder. The otherwise idle Jetson AGX
Orin runs in MAXN mode with \texttt{jetson\_clocks} enabled and GPU/EMC
frequencies fixed at 1300.5/3199\,MHz. We collect six measurements per route
in balanced order. Each runs in a fresh route-specific process with one
excluded warm-up, cleared VAE temporal state, and one CUDA-synchronized timed
decode.

TensorRT uses 99 fixed-shape learned-weight engines spanning 33 runtime
signatures and covers all 169 \texttt{WanCausalConv3d} invocations without
eager fallback. Engine construction, deserialization, and context creation
occur outside steady-state timing and are reported separately. Fresh processes
also prevent TensorRT residency from affecting eager or lowering memory
measurements.

Device telemetry records temperature, GPU and EMC frequencies, and available
power rails. The hottest sensor reaches 71.1$^\circ$C, with no observed
load-time GPU or EMC frequency reduction. Kernel logs nevertheless record
hot-surface cooling-state transitions, so we do not claim thermal throttling
is absent. Because no reliable whole-board \texttt{VDD\_IN} rail is exposed,
available power rails are reported independently rather than summed.

\subsection{Numerical Validation and Memory Accounting}

Same-latent evaluations reconstruct the VAE, clear temporal state, and verify
identical latent hashes between routes. The baseline uses the original Wan
causal Conv3D implementation, while strict mode terminates the lowering
evaluation on any unsupported route or exception. We report maximum and mean
absolute differences, RMSE, cosine similarity, and NaN/Inf mismatches in BF16
and FP32, with FP64 accumulation for cosine similarity.

A paired Edge generation uses the same checkpoint, image, prompt, seed,
scheduler, step count, and precision while capturing the encoder output,
pre-decode latent, and decoded tensor. This measures finite-precision
propagation through 50 unchanged denoising steps rather than isolated decoder
equivalence.

Peak anonymous RSS comes from \texttt{/proc/self/status}, while peak CUDA
allocated and reserved memory come from PyTorch's caching allocator. Because
the Jetson AGX Orin uses unified memory, these are overlapping views and are
not summed.

\section{Results}
\label{sec:results}

\subsection{Cosmos3-Edge Deployment}
\label{sec:edge-results}

\Cref{tab:edge-latency} presents the complete-pipeline timing decomposition.
At 25 frames, the lowering reduces median VAE-decoder latency from 90.05 to
13.03~s, corresponding to a $6.91\times$ speedup. Complete image-to-video
generation decreases from 180.68 to 81.94~s, producing a $2.21\times$
end-to-end speedup. The five-frame workload exhibits the same pattern, with
$7.92\times$ decoder acceleration and $1.85\times$ complete-generation
acceleration.

\begin{table}[!ht]
\caption{Cosmos3-Edge complete image-to-video latency decomposition. Values
are medians over three fresh-container runs per route.
$T_{\mathrm{dit+other}} =
T_{\mathrm{e2e}}-T_{\mathrm{encode}}-T_{\mathrm{decode}}$.}
\label{tab:edge-latency}
\centering
\scriptsize
\setlength{\tabcolsep}{5pt}
\begin{tabular*}{\textwidth}{@{\extracolsep{\fill}}clrrr@{}}
\toprule
Frames & Phase & Baseline (s) & Lowered (s) & Speedup \\
\midrule
5
& VAE encode
& 5.325
& 0.728
& $7.32\times$ \\

5
& Transformer + other
& 21.516
& 21.581
& $\approx1.00\times$ \\

5
& VAE decode
& 18.798
& 2.374
& $7.92\times$ \\

5
& Complete generation
& 45.639
& 24.681
& $1.85\times$ \\
\midrule
25
& VAE encode
& 25.561
& 4.024
& $6.35\times$ \\

25
& Transformer + other
& 65.072
& 65.142
& $\approx1.00\times$ \\

25
& VAE decode
& 90.046
& 13.027
& $6.91\times$ \\

25
& Complete generation
& 180.682
& 81.935
& $2.21\times$ \\
\bottomrule
\end{tabular*}
\end{table}

The patched Wan causal Conv3D class accelerates supported calls in both VAE
encoding and decoding, while transformer-and-other latency remains essentially
unchanged. At 25 frames, all 237 decoder calls use the fast path; the encoder
records 169 fast-path calls and 12 guarded fallbacks from two stride-two
temporal downsampling modules outside the unit-stride support contract.

\subsection{Decoder Sweeps and Architecture Transfer}
\label{sec:transfer-results}

\Cref{tab:transfer-latency} compares VAE-decoder latency across the three
evaluated systems. The repeated Cosmos3-Edge sweep achieves
$6.92\times$--$7.93\times$ acceleration over 5--49 frames, with run-to-run
standard deviation below 0.1~s. Cosmos3-Nano achieves
$4.74\times$--$6.65\times$ decoder acceleration, while LingBot-World achieves
$2.70\times$--$5.75\times$ using the unchanged lowering. Across the measured
decoder boundaries, all optimized \texttt{WanCausalConv3d} calls use the fast
path for all three systems, with no decoder fallbacks.

\begin{table}[!ht]
\caption{VAE-decoder BF16 latency. Cosmos3-Edge values are medians of $n=3$
runs per route; Cosmos3-Nano and LingBot-World values have $n=1$. Coverage
reports fast-path hits over \texttt{WanCausalConv3d} invocations within the
measured decoder boundary, excluding pre-target activity. Cosmos3-Nano records
1,083/1,083 fast-path executions across the sweep.}
\label{tab:transfer-latency}
\centering
\scriptsize
\setlength{\tabcolsep}{3pt}
\begin{tabular*}{\textwidth}{@{\extracolsep{\fill}}lcrrrc@{}}
\toprule
Model & Frames & C3D (s) & C2D (s) & Speedup & Coverage \\
\midrule
Cosmos3-Edge & 5  & 18.749  & 2.365  & 7.93$\times$ & 67/67 \\
Cosmos3-Edge & 9  & 33.095  & 4.458  & 7.42$\times$ & 101/101 \\
Cosmos3-Edge & 17 & 61.573  & 8.623  & 7.14$\times$ & 169/169 \\
Cosmos3-Edge & 33 & 118.519 & 16.976 & 6.98$\times$ & 305/305 \\
Cosmos3-Edge & 49 & 175.569 & 25.379 & 6.92$\times$ & 441/441 \\
\midrule
Cosmos3-Nano & 5  & 20.52  & 3.78  & 5.43$\times$ & 67/67 \\
Cosmos3-Nano & 9  & 33.44  & 7.05  & 4.74$\times$ & 101/101 \\
Cosmos3-Nano & 17 & 106.45 & 18.30 & 5.82$\times$ & 169/169 \\
Cosmos3-Nano & 33 & 253.26 & 38.08 & 6.65$\times$ & 305/305 \\
Cosmos3-Nano & 49 & 363.83 & 58.36 & 6.23$\times$ & 441/441 \\
\midrule
LingBot-World & 5  & 14.21  & 5.27  & 2.70$\times$ & 65/65 \\
LingBot-World & 9  & 26.74  & 5.29  & 5.05$\times$ & 98/98 \\
LingBot-World & 17 & 47.11  & 9.90  & 4.76$\times$ & 164/164 \\
LingBot-World & 33 & 102.24 & 17.77 & 5.75$\times$ & 296/296 \\
LingBot-World & 49 & 132.84 & 27.76 & 4.79$\times$ & 428/428 \\
\bottomrule
\end{tabular*}
\end{table}

All optimized Cosmos3-Edge and LingBot-World decoder calls use the fast path,
with no fallbacks, unsupported cases, or execution exceptions. Cosmos3-Nano
similarly records complete measured-decoder coverage, with 1,083/1,083
\texttt{WanCausalConv3d} invocations taking the fast path and no fallback,
unsupported, or exception increments across the sweep. Its complete-generation
acceleration ranges from $2.58\times$ to $3.30\times$.

The evaluations provide two complementary forms of evidence. Cosmos3-Edge and
Cosmos3-Nano use the byte-identical Wan2.2 VAE, but exercise it through
different generation modes and pipeline boundaries. LingBot-World instead
changes latent channels, decoder width, compression factors, feature-map
sizes, module count, and runtime shapes. Its acceleration therefore provides
the distinct-architecture transfer result.

\subsection{Numerical Agreement}
\label{sec:numerical-results}

\Cref{tab:numerics} presents same-latent BF16 agreement at 17 output frames.
Additional Cosmos3-Edge measurements across the complete frame sweep remain
stable: cosine similarity is at least 0.99998, maximum absolute difference is
at most 0.045, and RMSE ranges from $2.6\times10^{-3}$ to
$3.1\times10^{-3}$. No NaN or Inf mismatches are observed.

\begin{table}[!ht]
\caption{Same-latent BF16 agreement at 17 output frames. Cosmos3-Edge and
LingBot-World use synthetic decoder-compatible latents; Cosmos3-Nano uses a
saved generated latent.}
\label{tab:numerics}
\centering
\footnotesize
\setlength{\tabcolsep}{4pt}
\begin{tabular*}{\textwidth}{@{\extracolsep{\fill}}lccccc@{}}
\toprule
Model & Full max & Mean abs. & RMSE & Cosine & NaN/Inf \\
\midrule
Cosmos3-Edge
& $4.49\mathrm{e}{-2}$ & $1.84\mathrm{e}{-3}$
& $2.75\mathrm{e}{-3}$ & 0.99998334 & 0/0 \\
Cosmos3-Nano
& $3.3203\mathrm{e}{-2}$ & $1.8781\mathrm{e}{-3}$
& $2.8215\mathrm{e}{-3}$ & 0.99998309 & 0/0 \\
LingBot-World
& $2.3440\mathrm{e}{-2}$ & $1.5640\mathrm{e}{-3}$
& $2.3230\mathrm{e}{-3}$ & 0.99998402 & 0/0 \\
\bottomrule
\end{tabular*}
\end{table}

FP32 reduces the discrepancy further: Cosmos3-Edge reaches cosine similarity
0.99999999 with maximum absolute difference $1.2\times10^{-3}$, while
Cosmos3-Nano and LingBot-World reach 0.99999998, consistent with
finite-precision execution rather than a cache-indexing error.

TensorRT also agrees closely with eager BF16 execution: maximum absolute
difference 0.1016, RMSE $4.63\times10^{-3}$, cosine similarity 0.999953, and
no NaN or Inf values. Neither route is bit-exact.

In complete Edge generation, encoder, pre-decode, and decoded tensors reach
cosine similarities of 0.999977, 0.998841, and 0.999331. This captures
finite-precision propagation through 50 unchanged denoising steps rather than
isolated decoder equivalence.

\subsection{Memory Behavior and Completion}
\label{sec:memory-results}

The lowering trades memory for latency through folded inputs, accumulation,
and cached spatial weight slices. Across the Nano and LingBot sweeps, the
largest observed increases are 1.70/3.30/1.60~GiB and
7.66/4.60/7.62~GiB in anonymous RSS/CUDA allocated/CUDA reserved memory,
respectively. The optimized 25-frame Edge route reaches approximately
12.6~GB peak CUDA allocation without out-of-memory failure.

In the clean 17-frame Edge comparison, eager execution and our lowering use
5.37 and 7.31~GiB peak CUDA allocation, with process VmHWM of 8.23 and
9.75~GiB. Cached tap slices occupy 1.06~GB; major temporary buffers include
approximately 0.41~GB folded input, 0.20~GB accumulation, and 0.61~GB cache
concatenation. Their lifetimes overlap and therefore do not form an additive
peak decomposition.

Full TensorRT reaches 14.57~GiB peak CUDA allocation and 34.19~GiB process
VmHWM; its 99 execution contexts report 13.87~GB of context memory and the
serialized engines occupy 3.16~GB. All routes fit on the evaluated 64-GB Orin,
and Edge and LingBot also complete the 49-frame decoder configuration. Because
Jetson uses unified memory, process and CUDA memory metrics are overlapping
views and are not summed.

\subsection{Production-Runtime Comparison}
\label{sec:production_runtime}

We compare against a fully specialized BF16 TensorRT deployment on the same
17-frame Cosmos3-Edge decoder. Table~\ref{tab:production_runtime} summarizes
the clean-device results.

\begin{table}[!ht]
\centering
\caption{Clean-device 17-frame decoder comparison on Jetson AGX Orin.
Latency is the median over six runs; coverage denotes accelerated
\texttt{WanCausalConv3d} calls.}
\label{tab:production_runtime}
\small
\setlength{\tabcolsep}{6pt}
\begin{tabular}{@{}lccc@{}}
\toprule
Method & Latency (s) & Speedup & Coverage \\
\midrule
PyTorch eager & 61.72 & $1.00\times$ & native \\
Ours          &  9.23 & $6.69\times$ & 169/169 \\
TensorRT      & \textbf{6.78} & \textbf{$9.10\times$} & 169/169 \\
\bottomrule
\end{tabular}
\end{table}

TensorRT is $1.36\times$ faster than our lowering but requires 99 fixed-shape
engine specializations across 33 runtime signatures, 1637.6\,s of AOT
compilation, 3.16\,GB of serialized engines, and 7.58\,s of cold setup.

Using the clean-device medians, the fraction of the eager-to-TensorRT latency
reduction captured by our lowering is
\[
\frac{61.7179 - 9.2289}{61.7179 - 6.7848}
= 95.55\%.
\]
Thus, our lowering achieves most of the steady-state gain without per-module
or per-state AOT engine construction.

\section{Discussion}
\label{sec:discussion}

\subsection{What Edge Establishes and What Transfers}

Cosmos3-Edge provides the strongest deployment evidence because the lowering
remains effective inside a complete 50-step image-to-video pipeline. The
transformer-and-other portion remains essentially unchanged, so the
complete-generation improvement follows the accelerated VAE phases. The
larger end-to-end gain at 25 frames likewise reflects the greater baseline
latency spent inside the VAE.

Cosmos3-Edge and Cosmos3-Nano use the same byte-identical Wan2.2 VAE and
therefore demonstrate different execution boundaries rather than architectural
transfer. LingBot-World provides that transfer evidence through a Wan2.1 VAE
with different latent channels, compression, decoder width, feature hierarchy,
module count, and runtime shapes. The claim remains limited to the evaluated
Wan family and Diffusers causal-convolution implementation, where the shared
opportunity is short-temporal, large-spatial calls whose valid frames form
contiguous intervals suitable for batched Conv2D execution.

\subsection{Performance, Memory, and Numerical Implications}

The production-runtime comparison clarifies the deployment point targeted by
our method. Under aggressive fixed-model AOT specialization, TensorRT reaches
6.78\,s versus 9.23\,s for our lowering. However, our route captures 95.55\%
of the absolute eager-to-TensorRT latency reduction without constructing 99
module- and state-specific engines. The distinction is therefore not whether
specialized Conv3D can achieve higher steady-state throughput, but how much
specialization and memory residency are required to reach that operating
point.

This distinction is particularly relevant for embodied deployment, where
decoder latency is only one resource constraint. Perception, planning, control,
and other co-resident components must share the same device memory. In the
clean 17-frame comparison, our lowering uses 7.31~GiB peak CUDA allocation and
9.75~GiB process VmHWM, compared with 14.57~GiB and 34.19~GiB for the
full-residency TensorRT deployment. TensorRT therefore provides the lowest
measured decoder latency but leaves substantially less memory headroom for the
rest of an embodied-AI stack. Reducing this residency could require additional
engine management, such as selective loading or eviction, which we do not
evaluate. Our lowering instead occupies a lighter runtime point while retaining
most of the measured steady-state acceleration.

The remaining memory overhead is also potentially reducible. Cached spatial
weight slices account for approximately 1.06~GB, while folded inputs, cache
concatenation, and accumulation constitute the main temporary buffers. Their
overlapping lifetimes prevent an additive peak decomposition but suggest
opportunities for selective weight caching, buffer reuse, or streamed
accumulation.

We also retain the custom-kernel result as a systems-level caution. Selected
Triton kernels~\cite{tillet2019triton} improve individual operations in
isolation, yet an integrated 17-frame experiment increases VAE-decoding
latency from 16.01 to 66.81~s despite no route misses or emergency Conv3D
fallbacks. Without an integrated kernel-time breakdown, we do not attribute
the regression to a specific mechanism; instead, it reinforces that
operator-level gains must be validated at the complete execution boundary.

Finally, BF16 and FP32 results support numerical agreement for the tested
inputs but do not establish perceptual equivalence or invariance in downstream
control. The complete Edge comparison captures propagation of small BF16
differences through unchanged denoising. The smaller FP32 discrepancy is
consistent with finite-precision execution and accumulation order rather than
an observed cache-indexing error.

\subsection{Limitations}
\label{sec:limitations}

Cosmos3-Edge complete-generation and decoder sweeps use three fresh-container
repetitions per route, while the clean 17-frame runtime comparison uses six
balanced repetitions. Cosmos3-Nano and LingBot-World sweep points use one
fresh-process measurement and are reported without variance or
statistical-significance claims. All experiments use one 64-GB NVIDIA Jetson
AGX Orin, and fresh-process execution does not represent sustained or
concurrent serving.

Complete generation is evaluated for Edge and Nano, whereas LingBot-World uses
synthetic decoder-compatible latents. Edge and Nano also share a byte-identical
Wan2.2 VAE, so architectural transfer is established only through the evaluated
Wan2.1 LingBot decoder. Other cache APIs, Conv3D classes, hardware platforms,
resolutions, schedulers, and software stacks remain untested.

TensorRT is $1.36\times$ faster on the clean 17-frame workload, demonstrating
that our representation is not globally optimal. We do not exhaustively
evaluate other compiler-generated, vendor-specific, or manually optimized
Conv3D implementations, and relative performance may differ across workloads
and hardware.

The complete Edge numerical evaluation uses one image, prompt, seed, and
scheduler and does not establish perceptual equivalence, downstream-control
invariance, or closed-loop task performance. The runtime comparison also
records hot-surface cooling-state transitions despite no observed GPU or EMC
frequency reduction, so we do not claim thermal throttling is absent; no
reliable whole-board input-power rail was available for system-level energy
measurement.

Finally, the lowering remains restricted to supported cases including
five-dimensional inputs, unit temporal stride, and one convolution group.
Twelve stride-two encoder calls in the 25-frame Edge pipeline therefore remain
on native Conv3D, so zero-fallback claims apply to the decoder rather than the
complete pipeline.

\section{Conclusion}
\label{sec:conclusion}

We presented a cache-aware runtime lowering that evaluates supported causal
Conv3D calls as batched Conv2D operations without changing pretrained model
weights or temporal-cache semantics. Across the evaluated Wan-family decoders,
it provides substantial decoder acceleration and complete fast-path coverage,
including a $6.91\times$ decoder speedup and $2.21\times$ complete-generation
speedup on the primary 25-frame Cosmos3-Edge workload. A full-coverage
TensorRT comparison shows that aggressive fixed-model AOT specialization can
achieve higher steady-state throughput, while our lowering provides a
lighter-weight runtime operating point without per-module or per-state engine
construction. The same lowering also transfers to a distinct Wan2.1 VAE,
indicating that the optimization is not specific to a single decoder instance.
Numerical and route validation further show that the observed gains preserve
the intended cache behavior while maintaining strong finite-precision
agreement. Together, these results highlight runtime operator representation
as a practical lever for accelerating pretrained video world-model decoders on
embedded GPUs.

\bibliographystyle{splncs04}
\bibliography{main}

\end{document}